\documentclass{article}
\usepackage{ijcai26}

\usepackage{times}
\usepackage{soul}
\usepackage{url}
\usepackage[hidelinks]{hyperref}
\usepackage[T1]{fontenc}
\usepackage[utf8]{inputenc}
\usepackage[small]{caption}
\usepackage{graphicx}
\usepackage{amsmath}
\usepackage{amsthm}
\usepackage{booktabs}
\usepackage{algorithm}
\usepackage{algorithmic}
\usepackage[switch]{lineno}

\usepackage{xcolor}
\usepackage{pifont}

\usepackage[most]{tcolorbox}
\tcbuselibrary{skins,breakable,listings}
\usepackage{listings}

\usepackage{amsfonts}

\usepackage{cuted} 

\title{HeroBench: A Benchmark for Long-Horizon Planning and Structured Reasoning in Virtual Worlds}

\author{
Petr Anokhin$^{1,2}$
\and
Roman Khalikov$^2$
\and
Stefan Rebrikov$^{5,6}$
\and
Viktor Volkov$1$
\and
Artyom Sorokin$^{1}$
\and
Vincent Bissonnette$^7$\\
\affiliations
$^1$AXXX\\
$^2$Lomonosov Moscow State University\\
$^5$Higher School of Economics\\
$^6$Kurchatov Institute\\
$^7$Independent Researcher\\
}

\begin{document}

\maketitle

\begin{abstract}
Large language models (LLMs) perform well on step-by-step reasoning benchmarks such as mathematics and code generation, yet their ability to carry out robust long-horizon planning under realistic constraints remains insufficiently evaluated. Existing planning benchmarks often rely on abstract domains or interactive feedback, obscuring end-to-end planning failures and feasibility errors.
We introduce HeroBench, a benchmark for evaluating long-horizon, hierarchical planning and structured reasoning in a complex RPG-inspired virtual world. Tasks require models to select numerically feasible equipment, reason over multi-level crafting and resource dependencies, and execute hundreds to thousands of actions as a single end-to-end plan. HeroBench integrates symbolic planning, numeric combat simulation, spatial reasoning, and resource management, while supporting scalable difficulty and adversarial distractors.
HeroBench evaluates executable plans through simulation, enabling both success-based and fine-grained progress metrics, as well as detailed failure mode analysis. An evaluation of 25 state-of-the-art LLMs reveals large performance disparities rarely observed in conventional reasoning benchmarks. While reasoning models perform substantially better, no model reliably solves the hardest tasks, highlighting persistent challenges in long-horizon autonomous planning.

\end{abstract}

%

\section{Introduction}

\begin{figure}[t]
  \centering
  \includegraphics[width=1\linewidth]{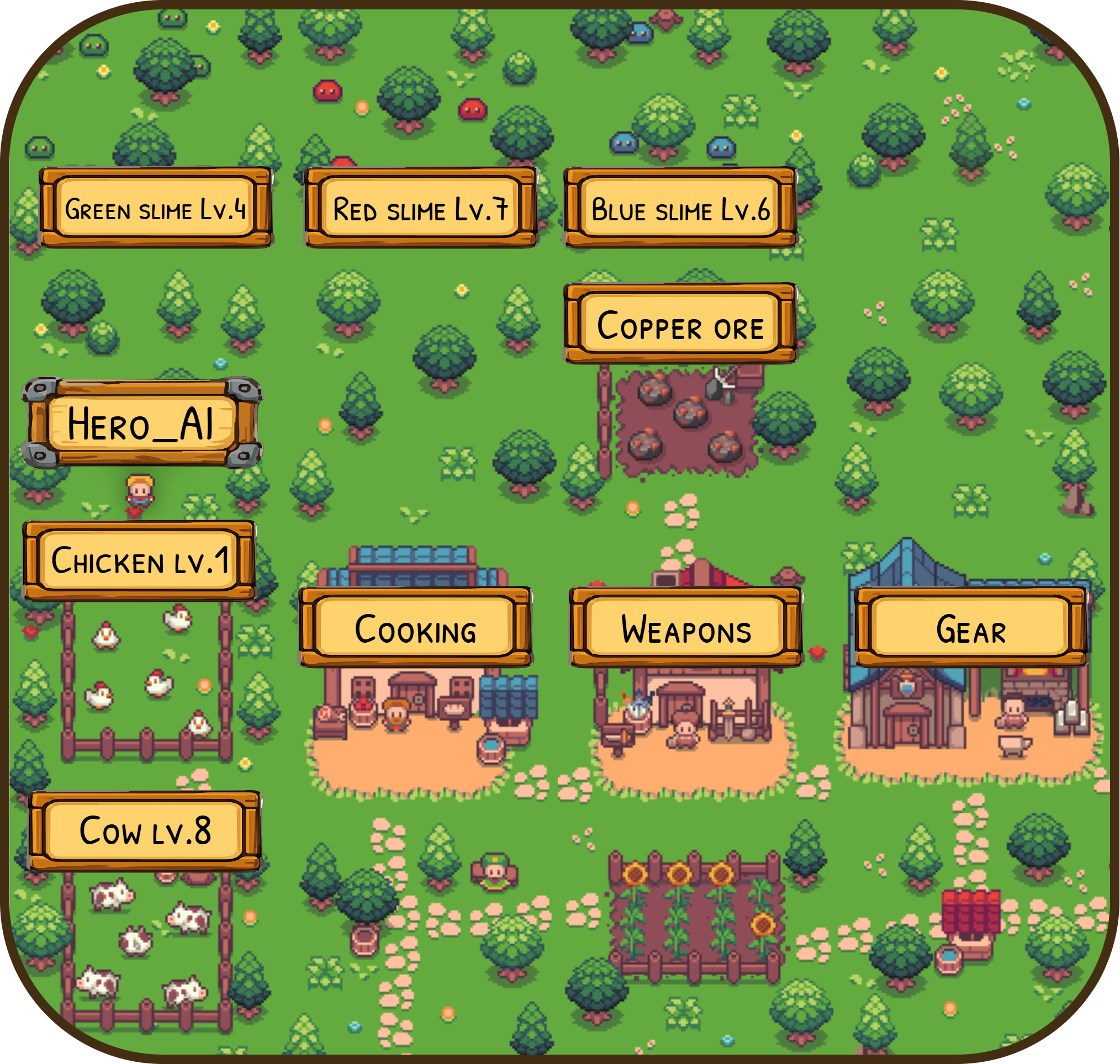}
  \caption{A section of the HeroBench virtual environment.
A grid-based RPG-inspired world where agents must navigate, gather resources, craft equipment, and defeat enemies. Each location encodes specific environmental elements, forming the foundation for generating complex, structured tasks that challenge long-horizon reasoning and planning abilities of language models.}
  \label{fig:env_example}
\end{figure}

\begin{figure}[t]
  \centering{
  \includegraphics[width=1\linewidth]{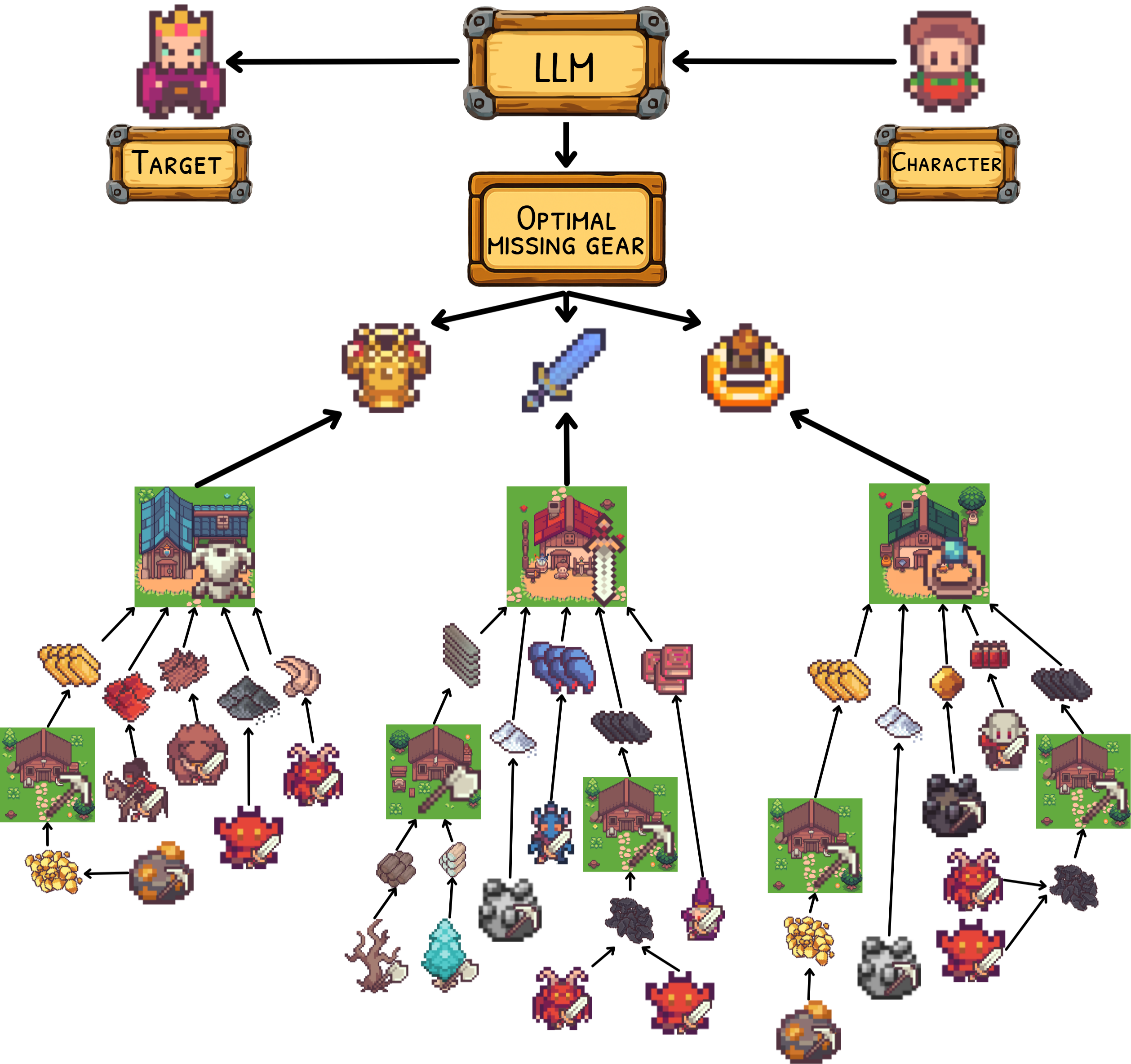}
  }
  \caption{Example of a task in the environment: the agent's ultimate goal is to defeat the target monster. To achieve this, agent must calculate the optimal gear by considering both its own and the monster's stats, and acquire all the necessary ingredients.}
  \label{fig:task_example}
\end{figure}

The rapid advancement of large language models (LLMs) has considerably expanded their applicability beyond traditional natural language processing tasks, establishing them as core components of autonomous agent systems across diverse domains. Today, LLM-driven agents are being developed to perform increasingly complex roles in automation and strategic decision-making. Central to these applications is the ability to perform robust long-term planning and reasoning, especially in scenarios that require the execution of action sequences over extended time horizons. Despite this progress, a growing body of research suggests that LLMs are not inherently capable of planning and often require external mechanisms to validate or supervise their plans \cite{kambhampati2024position,NEURIPS2023_efb2072a}.

Recent reinforcement learning (RL)-based training approaches have enhanced the reasoning capabilities of LLMs, giving rise to large reasoning models (LRM) such as OpenAI o1 and Deepseek R1 \cite{deepseekai2025deepseekr1incentivizingreasoningcapability}. These models are capable of producing extended chains of thought and can engage in partial self-verification of their outputs. This has led to notable performance gains, particularly in domains such as mathematics and programming - areas that are well-suited to formal verification and commonly included in model training. However, their performance remains suboptimal in tasks that require long-term planning \cite{valmeekam2025a,shojaee2025illusionthinkingunderstandingstrengths}.

Current claims regarding LLMs' planning abilities are often based on evaluations using standard algorithmic benchmarks, with tasks like Blocksworld being the primal example \cite{NEURIPS2023_7a92bcde}. Although such environments are easily scalable, they lack the complexity and variability characteristic of real-world scenarios. Other benchmarks that adopt game-based or real-world settings either fail to provide tasks with sufficient depth to challenge current models, or rely heavily on continuous environmental feedback. This reliance makes it difficult to isolate a model’s planning capabilities and to evaluate performance in domains where robust end-to-end planning is essential and failures are unacceptable. 

To address these limitations, we introduce HeroBench, a benchmark specifically designed to evaluate long-horizon, hierarchical planning under constraints. Built upon the ArtifactsMMO environment \cite{artifacts}, originally developed to assess programming skills via script-based gameplay, HeroBench features a classic RPG setting in which agents gather resources, craft equipment, and confront enemies. HeroBench simultaneously satisfies a set of properties that no existing planning benchmark covers in combination making it a strong test setup for LRMs (Table \ref{tab:benchmark_comparison}). Although HeroBench is framed as a game-based environment, it is designed to evaluate one-shot, end-to-end planning under strict feasibility constraints. Agents must commit to a complete plan before execution, with no opportunity for mid-course correction, closely mirroring domains in which errors are costly or irreversible. Tasks require hierarchical task decomposition, exact resource and dependency accounting, numeric feasibility reasoning, resistance to distractors, and sustained reasoning over long contexts. Crucially, HeroBench is designed not only to measure planning capability, but to assess the reliability of model behavior across long reasoning horizons.

We evaluated 25 state-of-the-art LLMs and LRMs, including open-source and proprietary models across different families and model sizes, revealing substantial performance differences.

\definecolor{yellow}{RGB}{230,160,0}
\definecolor{dark}{RGB}{0,150,0}

\begin{table*}[t]
\centering
\footnotesize
\caption{Comparison of Benchmarks Across Reasoning and Robustness Criteria}
\label{tab:benchmark_comparison}
\begin{tabular}{lccccccccc}
\hline
\textbf{Benchmark} 
& \textbf{One-shot} 
& \textbf{Max Act.} 
& \textbf{Feasibility} 
& \textbf{Analysis} 
& \textbf{Hierarch.} 
& \textbf{Numeric} 
& \textbf{Distract} 
& \textbf{Long-cont.} 
& \textbf{LRM} \\
\hline
PlanBench     
& \textcolor{yellow}{\ding{51}} 
& 47   
& \textcolor{red}{\ding{55}}
& \textcolor{yellow}{\ding{51}} 
& \textcolor{red}{\ding{55}} 
& \textcolor{red}{\ding{55}} 
& \textcolor{red}{\ding{55}} 
& \textcolor{red}{\ding{55}} 
& \textcolor{yellow}{\ding{51}} \\

PlanCraft      
& \textcolor{red}{\ding{55}} 
& 30   
& \textcolor{yellow}{\ding{51}} 
& \textcolor{yellow}{\ding{51}} 
& \textcolor{yellow}{\ding{51}} 
& \textcolor{yellow}{\ding{51}} 
& \textcolor{yellow}{\ding{51}} 
& \textcolor{yellow}{\ding{51}} 
& \textcolor{yellow}{\ding{51}} \\

Travel Planner
& \textcolor{red}{\ding{55}} 
& 30   
& \textcolor{yellow}{\ding{51}} 
& \textcolor{yellow}{\ding{51}} 
& \textcolor{yellow}{\ding{51}} 
& \textcolor{yellow}{\ding{51}} 
& \textcolor{red}{\ding{55}} 
& \textcolor{yellow}{\ding{51}} 
& \textcolor{yellow}{\ding{51}} \\

Natural Plan  
& \textcolor{yellow}{\ding{51}} 
& 19   
& \textcolor{yellow}{\ding{51}} 
& \textcolor{yellow}{\ding{51}} 
& \textcolor{yellow}{\ding{51}} 
& \textcolor{yellow}{\ding{51}} 
& \textcolor{red}{\ding{55}} 
& \textcolor{red}{\ding{55}} 
& \textcolor{red}{\ding{55}} \\

ScienceWorld  
& \textcolor{red}{\ding{55}} 
& 100  
& \textcolor{yellow}{\ding{51}} 
& \textcolor{yellow}{\ding{51}} 
& \textcolor{yellow}{\ding{51}} 
& \textcolor{yellow}{\ding{51}} 
& \textcolor{yellow}{\ding{51}} 
& \textcolor{yellow}{\ding{51}} 
& \textcolor{yellow}{\ding{51}} \\

Robotouille   
& \textcolor{yellow}{\ding{51}} 
& 82   
& \textcolor{yellow}{\ding{51}} 
& \textcolor{dark}{\ding{51}\ding{51}} 
& \textcolor{yellow}{\ding{51}} 
& \textcolor{yellow}{\ding{51}} 
& \textcolor{red}{\ding{55}} 
& \textcolor{yellow}{\ding{51}} 
& \textcolor{yellow}{\ding{51}} \\

ALFWorld      
& \textcolor{red}{\ding{55}} 
& 50   
& \textcolor{red}{\ding{55}}  
& \textcolor{yellow}{\ding{51}} 
& \textcolor{yellow}{\ding{51}} 
& \textcolor{red}{\ding{55}} 
& \textcolor{yellow}{\ding{51}} 
& \textcolor{red}{\ding{55}} 
& \textcolor{red}{\ding{55}} \\

LogiPlan      
& \textcolor{yellow}{\ding{51}} 
& 300  
& \textcolor{yellow}{\ding{51}} 
& \textcolor{yellow}{\ding{51}} 
& \textcolor{red}{\ding{55}} 
& \textcolor{red}{\ding{55}} 
& \textcolor{red}{\ding{55}} 
& \textcolor{yellow}{\ding{51}} 
& \textcolor{yellow}{\ding{51}} \\

HeroBench     
& \textcolor{yellow}{\ding{51}} 
& 1000+ 
& \textcolor{dark}{\ding{51}\ding{51}} 
& \textcolor{dark}{\ding{51}\ding{51}} 
& \textcolor{dark}{\ding{51}\ding{51}} 
& \textcolor{dark}{\ding{51}\ding{51}} 
& \textcolor{dark}{\ding{51}\ding{51}} 
& \textcolor{dark}{\ding{51}\ding{51}} 
& \textcolor{dark}{\ding{51}\ding{51}} \\
\hline
\end{tabular}
\caption{
\textit{One-shot}: end-to-end planning performed in a single generation;
\textit{Max Actions}: maximum number of environment actions required in the hardest tasks;
\textit{Feasibility}: requirement to satisfy multiple and diverse constraints;
\textit{Analysis}: availability of error analysis tools;
\textit{Hierarchical}: presence of deep hierarchical task structures;
\textit{Numeric}: requirement of numeric reasoning in addition to symbolic planning;
\textit{Distractors}: inclusion of distractor conditions;
\textit{Long-context}: requirement to process long, detailed environment descriptions;
\textit{LRM}: difficulty level appropriate for evaluating state-of-the-art large reasoning models.
\textcolor{dark}{\ding{51}\ding{51}} strong presence,
\textcolor{yellow}{\ding{51}} partial presence,
\textcolor{red}{\ding{55}} absence.
}
\end{table*}

\section{HeroBench}
\subsection{Dataset description}

The environment displayed in Fig.~\ref{fig:env_example} is a structured RPG-style game with a discrete action space. The world is organized as a grid of 70 locations, each containing specific elements such as resource nodes, workshops, or monster spawns. The environment includes 25 distinct monsters, 17 resource types for crafting, and 208 unique items, including gear and crafting components. All environment data is defined through JSON files. 

The dataset consists of tasks of varying difficulty levels, each requiring the player to defeat a specific monster in the game or craft an item. An example prompt is provided in the Appendix \ref{app:prompt}. The player starts with a character of appropriate level and a specific set of equipment. The tasks are divided into two categories: purely crafting tasks, which do not require any combat, and tasks that involve defeating enemies. In the latter case, defeating a monster typically requires the character to craft one or more additional items beforehand. The difficulty of a task is determined by the number of required items and the number of steps involved in crafting them. For example, crafting a simple bronze sword may require only mining and smelting ore, whereas crafting a high-level item can involve many steps, including obtaining drops from defeated monsters and gathering and refining multiple types of resources (Fig.~\ref{fig:task_example}).


\subsection{Task Formalization}
\label{formalism}
A task instance induces a deterministic planning problem with state
$s=\langle \ell, I, E, L, H\rangle$, where $\ell\in\mathbb{Z}^2$ is the agent location,
$I$ is an inventory multiset, $E$ is an equipment assignment (one item per slot),
$L$ are profession skill levels (e.g., mining, woodcutting, smithing), and $H$ denotes derived combat stats
(HP, attacks, resistances, damage amplifiers) computed from the base character plus effects of $E$.
We write $G(E)$ for the (slot-wise) set of equipped items implied by $E$.

Actions are deterministic and have location- and inventory-dependent preconditions:
$\texttt{move}(x,y)$ updates $\ell$; $\texttt{gather}$ yields one unit of the resource available at $\ell$;
$\texttt{fight}(m)$ is applicable only if monster $m$ is co-located and yields guaranteed drops upon victory;
$\texttt{equip}/\texttt{unequip}$ updates $E$ subject to slot constraints.
Crafting is captured by $\texttt{craft}(i,q)$, applicable only if (i) $\ell$ contains the required station
$\sigma(i)$, (ii) $L$ satisfies a skill threshold $\tau(i)$, and (iii) $I$ contains the required inputs.

\paragraph{Item--ingredient--resource hierarchy.}
Each craftable item $i$ defines a recipe hyperedge
$R_i=\{(j,n_{ij})\}$, consuming $n_{ij}$ units of ingredient $j$.
Recipe expansion induces a dependency hypergraph/DAG
\[
\begin{aligned}
\text{target item/gear}
&\rightarrow \text{ingredients} \\
&\rightarrow \text{crafted sub-ingredients} \\
&\rightarrow (\text{gather nodes} \cup \text{monster drops}) .
\end{aligned}
\]
where leaves are primitive acquisition operators (gather/fight) and internal nodes are
craft operations constrained by station locations and skill thresholds. Exact planning
requires propagating multiplicities through the hierarchy to compute total required counts.

\paragraph{Embedded numeric feasibility and gear selection.}
Tasks with a combat goal ``defeat monster $m$'' introduce a numeric feasibility constraint under a turn-based,
type-specific combat simulator. For any chosen equipment assignment $E$, the agent computes derived stats
$H(E)$ and verifies $\textsc{Win}(m;H(E))$.
Damage per turn is computed for each non-zero damage type $t$:
\[
\begin{aligned}
d^{\text{char}}_t
&= a_t\Bigl(1+\frac{b_t}{100}\Bigr)
   - a_t\Bigl(\frac{r^{m}_t}{100}\Bigr), \\
d^{m}_t
&= a^{m}_t
   - a^{m}_t\Bigl(\frac{r^{\text{char}}_t}{100}\Bigr).
\end{aligned}
\]
where $a_t$ is base attack of the attacker in type $t$, $b_t$ is matching damage amplification,
and $r_t$ is the defender's resistance in type $t$ (allowing negative values).
Total per-turn damage is $\sum_t d_t$, and the simulated fight is feasible iff the monster's HP reaches
$0$ before the character's HP.

\paragraph{Sources of complexity.}
HeroBench combines: (i) hierarchical resource planning via multi-level recipe expansion with exact
counting, (ii) spatial planning to route between resource nodes, monsters, and multiple crafting stations,
and (iii) discrete optimization over equipment assignments $E$ (slot-wise) coupled to numeric combat
feasibility. Difficulty arises from the tight interaction between combinatorial choices (gear and ordering)
and numeric constraints (amplifiers/resistances) that determine whether a plan can satisfy the combat goal.


\begin{figure*}[ht]
  \centering
  \includegraphics[width=1.0\textwidth]{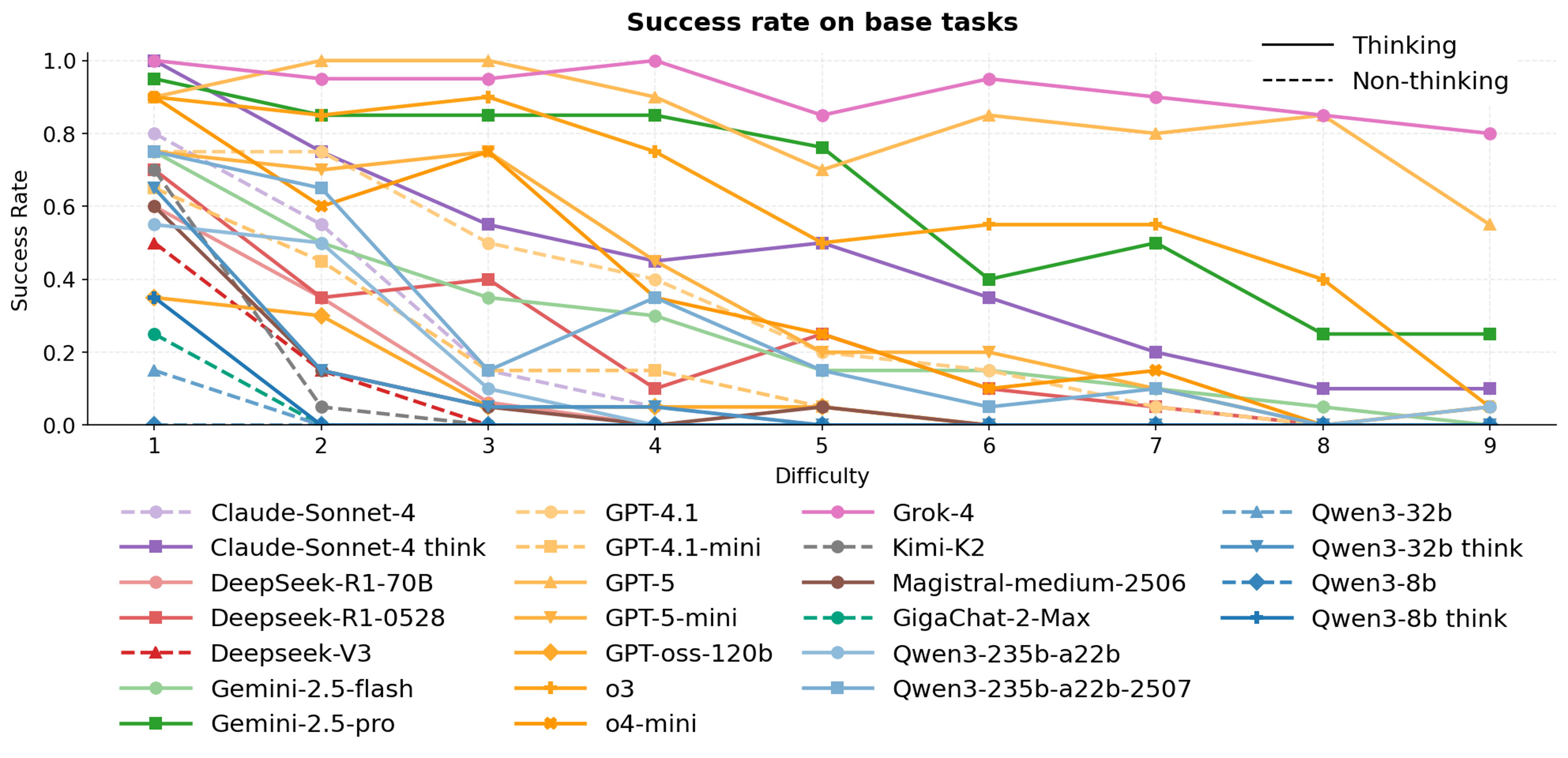}
  \caption{Success rate of LLMs across nine base-task difficulty levels. Solid lines correspond to reasoning (thinking) models, while dashed lines represent standard (non-thinking) variants.}
  \label{fig:results_difficulties}
\end{figure*}

\section{Benchmark Task Generation Pipeline}
\label{sec:task_generation}

We present a systematic pipeline for constructing benchmark tasks for HeroBench. Each generated instance
is a planning problem of the form in Sec.~\ref{formalism}, with (optional) combat feasibility
$\textsc{Win}(m;H(E))$ and a crafting/resource dependency DAG induced by recipes.

\paragraph{Monster initialization.}
Let $\mathcal{M}=\{m_1,m_2,\dots,m_N\}$ denote the set of all monsters in the game.
To generate a new combat task, we first sample a target monster
$m \sim P_{\mathcal{M}}(\cdot \mid d_{\mathrm{target}})$ conditioned on the desired difficulty.
Each monster $m$ is associated with a statistics vector $\mathbf{m}$ (health, attacks, resistances, etc.)
and a combat difficulty level $L(m)$, defined as the minimal \emph{character combat level} under which
$m$ can be defeated under standard conditions.

\paragraph{Combat simulation / feasibility check.}
For a fixed monster $m$ and an equipment assignment $E$, combat is simulated in a turn-based manner
from the derived stats $H(E)$ and monster stats $\mathbf{m}$.
We write the simulator outcome as the feasibility predicate
\[
\textsc{Win}(m;H(E)) \in \{0,1\},
\]
which returns $1$ iff the character defeats $m$.

\paragraph{Minimal winning equipment search.}
Let $\mathcal{I}_{\leq L(m)}$ be the set of equipment items whose level requirement is at most $L(m)$.
We search over feasible equipment assignments $E$ using items from $\mathcal{I}_{\leq L(m)}$ and define
a \emph{minimal winning equipment assignment} $E^*$ as one that satisfies:
\begin{itemize}
    \item $\textsc{Win}(m;H(E^*)) = 1$;
    \item for any strict ablation of equipped items (i.e., any $E'$ obtained by unequipping at least one item from $E^*$ while respecting slot constraints), $\textsc{Win}(m;H(E'))=0$.
\end{itemize}
Equivalently, $E^*$ is minimal in the sense that removing any single equipped item causes failure.

From $E^*$ we induce the equipped item set $G(E^*)$ and partition it into:
\begin{itemize}
    \item {Initially equipped items} ($\mathcal{I}_{\mathrm{eq}}$): items present in $E$ at episode start,
    \item {Missing items} ($\mathcal{I}_{\mathrm{miss}}$): items that must be acquired/crafted during the episode.
\end{itemize}
The \emph{base} task difficulty is $d = |\mathcal{I}_{\mathrm{miss}}|$, and is refined using acquisition and crafting costs.

\paragraph{Crafting and environment analysis.}
For each item $i \in \mathcal{I}_{\mathrm{miss}}$, we traverse the recipe dependency DAG induced by $\{R_i\}$
(Sec.~\ref{formalism}) to extract all required materials, intermediate crafted components,
relevant monsters (for drops), and required locations (resource nodes and stations).
The total task difficulty is:
\[
D_{\mathrm{total}} = |\mathcal{I}_{\mathrm{miss}}| + \sum_{i \in \mathcal{I}_{\mathrm{miss}}} \mathrm{cost}(i),
\]
where $\mathrm{cost}(i)$ is a crafting/acquisition cost function (App.~A).

\paragraph{Auxiliary item validation.}
To enforce robust solution paths, we compute an auxiliary set of valid items $\mathcal{I}_{\mathrm{aux}}$
such that:
\begin{itemize}
    \item the character equipped with $\mathcal{I}_{\mathrm{eq}} \cup \mathcal{I}_{\mathrm{aux}}$ can defeat all non-target monsters present in the scenario;
    \item but cannot defeat the target monster $m$ without acquiring all items in $\mathcal{I}_{\mathrm{miss}}$.
\end{itemize}
Formally, for any monster $m_j$ with stats $\mathbf{m}_j$ present in the scenario,
\[
\textsc{Win}(m_j; H(E_{\mathrm{eq}\cup\mathrm{aux}})) = \mathbb{I}[m_j \neq m],
\]
where $E_{\mathrm{eq}\cup\mathrm{aux}}$ denotes any equipment assignment consistent with equipping
$\mathcal{I}_{\mathrm{eq}} \cup \mathcal{I}_{\mathrm{aux}}$ (subject to slot constraints).

Crafting-only tasks follow the same pipeline but omit the combat feasibility steps and directly specify goal items,
with difficulty determined solely by traversal of the crafting dependency DAG.

\paragraph{Leveling mechanics.}
In the base task set, the profession skill-level vector $L$ is initialized to match the highest requirement
among items in $\mathcal{I}_{\mathrm{miss}}$.
In the extended version, we introduce skill progression: the agent starts at level 1 in all relevant professions.
To support skill growth, the environment description includes accessible resource nodes and associated experience rewards,
so that crafting high-level items requires planning a sequence of actions that incrementally increases $L$.

\paragraph{Noise item injection.}
To increase task complexity and test robustness, we optionally add noise items $\mathcal{I}_{\mathrm{noise}}$:
plausible, high-level gear items that appear valid based on stats and level filters, but are impossible to craft
because at least one required prerequisite (ingredient/resource/station) is omitted from the environment.

\begin{table}[ht]
\centering
\small
\setlength{\tabcolsep}{4pt} 
\begin{tabular}{c c c c}
\hline
\textbf{Difficulty} 
& \begin{tabular}{c}
Environmental \\
actions
\end{tabular}
& \begin{tabular}{c}
Crafting \\
complexity
\end{tabular}
& \begin{tabular}{c}
Unique \\
milestones
\end{tabular} \\
\hline
1  & $40 \pm 31$   & $8 \pm 3$   & $6 \pm 3$ \\
2  & $122 \pm 52$  & $18 \pm 3$  & $12 \pm 3$ \\
3  & $178 \pm 53$  & $28 \pm 3$  & $17 \pm 3$ \\
4  & $261 \pm 51$  & $38 \pm 3$  & $21 \pm 3$ \\
5  & $304 \pm 80$  & $48 \pm 3$  & $26 \pm 3$ \\
6  & $396 \pm 70$  & $58 \pm 3$  & $29 \pm 4$ \\
7  & $461 \pm 121$ & $68 \pm 3$  & $32 \pm 4$ \\
8  & $552 \pm 161$ & $78 \pm 3$  & $37 \pm 3$ \\
9  & $663 \pm 143$ & $88 \pm 3$  & $39 \pm 3$ \\
10 & $937 \pm 242$ & $88 \pm 3$  & $41 \pm 3$ \\
\hline
\end{tabular}
\caption{Difficulty levels in HeroBench. All values are reported as (mean $\pm$ SD). \textit{Environmental actions} denote the number of actions in the environment required to complete the task. \textit{Crafting complexity} represents the number and depth of crafting steps involved, computed as 2 points for each craftable item and 1 point for each non-craftable ingredient. \textit{Unique milestones} indicate the number of distinct subtasks that must be completed, such as defeating a monster or obtaining a specific item.}
\label{tab:difficulty_summary}
\end{table}

\paragraph{Task Representation}

Each task is serialized as a structured JSON object, specifying: target monster or craft item name; equipped and missing items; full character state; environment information (dependencies, required monsters, locations, etc.).
Prompts for language models are generated from these objects to ensure reproducibility.
The final dataset contains 844 tasks with difficulty levels ranging from 2 to 97. Input prompt lengths vary from 1k to 11k tokens. For our experiments, we selected a subset of 180 tasks, divided into 9 difficulty brackets. Leveling and noise mechanics can be incorporated on top of the base tasks. The benchmark is highly accessible: the environment is fully defined in a single prompt, and all scripts needed to score the agent’s plan are provided.

\paragraph{Evaluation}

The LLM or an agentic system is prompted to generate a sequence of  actions in Python that solves the given task. The generated sequence is then parsed and executed in the environment, with the resulting simulation logs recorded for analysis.

Two evaluation metrics are used: \textit{Success}, indicating whether the final goal (crafting the target item or defeating the target monster) is achieved; and \textit{Progress score}, which reflects partial completion based on valid intermediate actions such as gathering, recycling, defeating required monsters, crafting, and equipping gear.

This dual-metric evaluation enables both binary assessment of task completion and fine-grained measurement of the agent's progress and problem-solving efficiency. We also provide an evaluation pipeline that offers comprehensive statistics on the types of errors made by the agents. These include mistakes in high level plan decomposition and optimal gear calculation, failures in determining the required amount of resources or appropriate level for item crafting, incorrect usage of provided information such as location coordinates, and improper code formatting in the response. This allows for a more precise assessment of the models' weaknesses.

\paragraph{Agentic systems}
We also developed and evaluated handcrafted agentic pipelines on HeroBench. Implementation details are provided in Appendix~\ref{app:multi_agent_descr}.

\section{Results}
We evaluated a wide range of state-of-the-art LLMs on our benchmark, encompassing both standard and reasoning models. Experiments were conducted using local FP16 Qwen-8B and Qwen-32B models, the OpenAI API for the o3 and o4-mini, and the OpenRouter API for the remaining models. All hyperparameters were set to default values. The reasoning budget was set to 'high' for OpenAI models and capped at 40,000 reasoning tokens for the other models, which was not exceeded in our experiments. All models were tested on the base set of tasks, while the top-performing reasoning models were further evaluated on harder difficulty levels incorporating additional mechanics.

\begingroup
\setlength{\tabcolsep}{3pt} 
\begin{table}[t]
\begin{small}
\centering
\begin{tabular}{@{}l@{\hspace{0.7em}}ccc@{}}
\toprule
\textbf{Model} & \textbf{Success \%} & \textbf{Score} & \textbf{Tokens} \\
\midrule
Qwen3 8b                          & 0.0  & 11.5 $\pm$ 6.8   & 2883 $\pm$ 1965 \\
Qwen3 32b                         & 1.7  & 21.9 $\pm$ 12.8  & 2074 $\pm$ 1222 \\
GigaChat 2 Max                    & 2.8  & 21.3 $\pm$ 15.4  & 1190 $\pm$ 228  \\
Qwen3 8b (t)                      & 3.9  & 28.8 $\pm$ 15.5  & 9680 $\pm$ 1224 \\
Deepseek-v3                       & 7.2  & 32.7 $\pm$ 17.9  & 1586 $\pm$ 430  \\
Kimi-K2                           & 8.3  & 29.6 $\pm$ 16.4  & 1309 $\pm$ 237  \\
GPT-oss-120b                      & 8.9  & 27.0 $\pm$ 8.7   & 9372 $\pm$ 2959 \\
Magistral-medium                  & 9.4  & 25.0 $\pm$ 18.8  & 10885 $\pm$ 1667 \\
Qwen3 32b (t)                     & 10.0 & 44.8 $\pm$ 17.1  & 9107 $\pm$ 1458 \\
DeepSeek-R1-70B                   & 11.2 & 27.5 $\pm$ 21.2  & 7448 $\pm$ 1029 \\
Qwen3-235b                        & 13.3 & 34.9 $\pm$ 20.5  & 12006 $\pm$ 1746 \\
GPT-4.1-mini                      & 16.1 & 53.9 $\pm$ 17.6  & 4555 $\pm$ 1398 \\
Claude-Sonnet-4                   & 17.2 & 50.6 $\pm$ 21.0  & 1578 $\pm$ 306  \\
Qwen3-235b-2507                   & 24.4 & 49.4 $\pm$ 18.6  & 11387 $\pm$ 2702 \\
Deepseek-R1-0528                  & 21.7 & 48.7 $\pm$ 22.5  & 10711 $\pm$ 2088 \\
Gemini-2.5-flash                  & 26.1 & 64.8 $\pm$ 13.7  & 11028 $\pm$ 4010 \\
GPT-4.1                           & 31.7 & 73.7 $\pm$ 10.3  & 3518 $\pm$ 1202 \\
o4-mini                           & 35.0 & 56.1 $\pm$ 23.5  & 21993 $\pm$ 8181 \\
GPT-5-mini                        & 35.0 & 59.8 $\pm$ 22.5  & 14126 $\pm$ 4169 \\
Claude-Sonnet-4 (t)               & 44.4 & 73.8 $\pm$ 16.9  & 16397 $\pm$ 4313 \\
o3                                & 60.6 & 84.6 $\pm$ 8.5   & 13897 $\pm$ 5250 \\
Gemini-2.5-pro                    & 62.9 & 86.6 $\pm$ 10.4  & 12935 $\pm$ 4295 \\
GPT-5                             & 83.9 & 95.0 $\pm$ 3.3   & 17851 $\pm$ 7149 \\
Grok-4                            & 91.7 & 95.3 $\pm$ 3.3   & 15470 $\pm$ 5838 \\
\bottomrule
\end{tabular}
\caption{Mean performance of all evaluated models across nine base task difficulty levels in HeroBench.
Columns show success rate (\%), score (mean $\pm$ SD), and tokens (mean $\pm$ SD). SD is computed across the nine difficulty-level averages for each model. Thinking-enabled variants are denoted by (t).}
\label{tab:model_performance}
\end{small}
\end{table}
\endgroup

\begin{table*}[t]
\begin{small}
\centering
\begin{tabular}{lrr|rrrrr}
\toprule
\textbf{Model} &
\multicolumn{2}{c|}{\textbf{Errors (mean $\pm$ SD)}} &
\multicolumn{5}{c}{\textbf{Failure Types (\% of all tasks)}} \\
& \textbf{High-level} & \textbf{Execution} 
& \textbf{Failed} & \textbf{Only Gear} & \textbf{Gear+Exec} & \textbf{Only Exec} & \textbf{Invalid Output} \\
\midrule
Qwen3 8b                  & $3.32 \pm 2.23$ & $2.59 \pm 0.49$ & 100.0 &  8.9 & 67.8 & 10.6 & 12.8 \\
Qwen3 32b                 & $3.62 \pm 2.33$ & $6.44 \pm 5.86$ &  98.3 &  5.0 & 77.2 &  7.2 &  7.2 \\
GigaChat 2 Max            & $3.89 \pm 2.49$ & $1.93 \pm 1.11$ &  97.2 & 17.2 & 61.1 &  5.6 & 11.7 \\
Qwen3 8b (think)          & $3.52 \pm 2.31$ & $2.69 \pm 0.88$ &  96.1 &  5.6 & 78.9 &  7.8 &  2.8 \\
Deepseek-v3               & $3.69 \pm 2.38$ & $1.39 \pm 0.63$ &  92.8 & 16.7 & 68.3 &  5.0 &  1.7 \\
Kimi-K2                   & $3.78 \pm 2.52$ & $1.34 \pm 0.71$ &  91.7 & 17.8 & 53.9 &  1.1 & 18.3 \\
GPT-oss-120b (think)             & $3.30 \pm 2.22$ & $4.90 \pm 5.95$ &  91.9 & 12.2 & 34.4 &  3.9 & 40.6 \\
Magistral-medium-2506 (think) & $3.10 \pm 2.48$ & $2.14 \pm 2.10$ &  90.6 & 16.1 & 32.8 &  2.8 & 36.7 \\
Qwen3 32b (think)         & $3.21 \pm 2.18$ & $4.89 \pm 4.09$ &  90.0 &  9.4 & 70.6 &  5.6 &  3.3 \\
DeepSeek-R1-70B (think)   & $3.25 \pm 2.64$ & $1.86 \pm 0.88$ &  86.7 & 11.1 & 41.7 &  6.7 & 26.7 \\
Qwen3-235b-a22 (think)    & $2.79 \pm 2.40$ & $2.79 \pm 1.91$ &  87.8 &  9.4 & 45.0 &  2.8 & 27.2 \\
GPT-4.1-mini              & $2.77 \pm 2.42$ & $1.91 \pm 1.79$ &  83.9 &  7.2 & 55.0 &  6.7 & 11.1 \\
Claude-Sonnet-4           & $3.41 \pm 2.59$ & $4.93 \pm 5.01$ &  82.8 &  0.6 & 60.6 &  3.3 & 16.1 \\
Qwen3-235b-a22-2507 (think)& $2.76 \pm 2.41$ & $1.89 \pm 1.15$ &  77.8 & 13.9 & 36.7 &  2.2 & 22.2 \\
Deepseek-R1-0528 (think)  & $2.73 \pm 2.46$ & $1.38 \pm 1.39$ &  78.3 & 19.4 & 39.4 &  4.4 & 12.2 \\
Gemini-2.5-flash (think)  & $1.29 \pm 1.27$ & $1.56 \pm 1.66$ &  73.9 & 17.8 & 30.6 &  7.8 & 10.6 \\
GPT-4.1                   & $1.50 \pm 1.39$ & $0.78 \pm 0.89$ &  68.3 & 26.1 & 32.2 &  3.3 &  2.2 \\
o4-mini (think)           & $2.12 \pm 2.11$ & $0.44 \pm 0.30$ &  65.0 & 22.8 & 22.2 &  3.9 & 14.4 \\
GPT-5-mini (think)        & $1.86 \pm 2.07$ & $0.37 \pm 0.32$ &  65.0 & 29.4 & 15.6 &  2.8 & 15.0 \\
Claude-sonnet-4 (think)   & $1.58 \pm 1.58$ & $0.34 \pm 0.21$ &  55.6 & 24.4 & 21.7 &  3.9 &  3.3 \\
o3 (think)                & $0.82 \pm 0.98$ & $0.18 \pm 0.17$ &  39.4 & 21.7 &  7.8 &  7.8 &  1.1 \\
Gemini-2.5-pro (think)    & $0.83 \pm 1.09$ & $0.10 \pm 0.12$ &  37.8 & 22.2 &  7.8 &  1.1 &  1.7 \\
GPT-5 (think)             & $0.19 \pm 0.28$ & $0.08 \pm 0.07$ &  16.1 &  8.3 &  1.1 &  5.6 &  0.6 \\
Grok-4 (think)            & $0.11 \pm 0.16$ & $0.02 \pm 0.04$ &   8.3 &  2.8 &  1.7 &  0.6 &  2.8 \\
\bottomrule
\end{tabular}
\caption{
Breakdown of failure types across models. First two columns show the mean $\pm$ SD number of errors per task for tasks with valid generated code. The first column reports the average number of items missing in the high-level plan, while the second reports the average number of low-level execution mistakes. The remaining columns show the percentage of tasks that failed due to: gear selection only, gear selection plus execution errors, execution-only errors, or invalid code. 
}
\label{tab:error_breakdown}
\end{small}
\end{table*}

\subsection{Base tasks}
The success rate over different difficulties is shown in Fig. \ref{fig:results_difficulties} and mean metrics over all difficulties are presented in table \ref{tab:model_performance}. Reasoning models consistently outperform standard models across all levels of task difficulty. However, the accuracy of most LRMs declines as complexity increases. In contrast to conventional mathematical and coding benchmarks, where open-source models approach the performance of leading proprietary ones, our evaluation reveals substantial variability in model performance. Notably, Grok 4 achieved the highest scores and exhibited the least performance degradation as task difficulty increased, clearly outperforming other models at higher difficulty levels.

Among non-reasoning models, GPT-4.1 demonstrated the best performance, outperforming several open-source reasoning models and achieving a success rate close to that of o4-mini, while even surpassing it in score. Its success rate was nearly double that of Claude Sonnet-4 (non-thinking), although it used more than twice the number of tokens to solve the tasks. Overall, GPT-4.1 and Claude Sonnet-4 (non-thinking) exhibited the best performance in terms of success per tokens spent.

We tested the Qwen3-235b-a22 and Qwen3-235b-a22-2507 models following a switch from the widely used GRPO \cite{shao2024deepseekmathpushinglimitsmathematical} algorithm to GSPO \cite{zheng2025groupsequencepolicyoptimization}, which is better suited for RL training of MoE architectures. This change resulted in a significant improvement in the models' planning capabilities, though still not sufficient to match the performance of proprietary models. 

Additional charts, including token usage efficiency, are presented in Appendix \ref{app:additional_results}.

\begin{table*}[t]
\centering
\small 
\setlength{\tabcolsep}{3pt} 
\renewcommand{\arraystretch}{1.2} 
\begin{tabular}{l
  |ccc
  |ccc
  |ccc}
\toprule
\textbf{Model} 
& \multicolumn{3}{c|}{\textbf{Base}} 
& \multicolumn{3}{c|}{\textbf{Leveling}} 
& \multicolumn{3}{c}{\textbf{Leveling+Noise}} \\
\cline{2-10}
& Succ (\%) & Score & Tokens 
& Succ (\%) & Score & Tokens 
& Succ (\%) & Score & Tokens \\
\midrule
o3 
    & 5  & 66.2 $\pm$ 32.1 & 20688 $\pm$ 2791
    & 0  & 26.6 $\pm$ 28.4 & 22606 $\pm$ 2788
    & 0  & 15.9 $\pm$ 12.0 & 23562 $\pm$ 3996 \\
Claude-Sonnet-4 
    & 10 & 42.6 $\pm$ 36.3 & 21366 $\pm$ 6036
    & 0  & 25.6 $\pm$ 19.0 & 24588 $\pm$ 6651
    & 0  & 21.9 $\pm$ 14.2 & 25404 $\pm$ 5697 \\
Gemini-2.5-pro 
    & 25 & 66.1 $\pm$ 26.6 & 18636 $\pm$ 3835
    & 10 & 32.7 $\pm$ 26.4 & 20047 $\pm$ 3141
    & 5  & 36.0 $\pm$ 28.5 & 21741 $\pm$ 3127 \\
GPT-5 
    & 55 & 90.6 $\pm$ 16.5 & 28052 $\pm$ 3776
    & 15 & 62.3 $\pm$ 32.6 & 31704 $\pm$ 3656
    & 20 & 59.9 $\pm$ 34.2 & 36052 $\pm$ 4196 \\
Grok-4 
    & 80 & 95.5 $\pm$ 14.2 & 22850 $\pm$ 4587
    & 65 & 92.9 $\pm$ 16.5 & 28361 $\pm$ 5953
    & 65 & 78.8 $\pm$ 31.8 & 33305 $\pm$ 6672 \\
\bottomrule
\end{tabular}
\caption{ 
Evaluation of five leading reasoning models under increased task complexity (difficulty 10). Results are shown for three conditions: \textit{Base} (standard level 9 tasks), \textit{Leveling} (requires skill progression before crafting), and \textit{Leveling+Noise} (adds adversarial distractor items). Metrics include success rate, progress score (mean $\pm$ SD), and token usage (mean $\pm$ SD).}
\label{tab:level_noise}
\end{table*}

\paragraph{Results analysis}
To understand the weaknesses and failure modes of various models, we provide a script for comprehensive analytics of model's performance on the tasks. It scores how many tasks failed due to errors in the high-level plan for selecting optimal gear, how many items were incorrectly chosen for the optimal outcome, how many mistakes the model made in executing the plan (e.g., incorrect amounts of resources, misusing environmental information, redundant steps, etc.), and how many plans failed due to incorrectly formatted output.


An analysis of the failure modes in Table \ref{tab:error_breakdown} reveals a clear bottleneck in high-level planning relative to low-level execution across model architectures. Although top-tier reasoning models such as Gemini 2.5 Pro, Grok-4, and GPT-5 exhibit substantially fewer execution errors, demonstrating their ability to reliably interpret and use contextual information, they continue to struggle with high-level search for optimal gear configurations. This challenge, which involves complex numeric reasoning and constraint satisfaction, indicates that determining optimal strategies remains more difficult than correctly implementing them. In contrast, smaller and open-source models fail more pervasively, often making simultaneous errors in both gear selection and execution. Other models, such as GPT-oss-120b, are further hindered by severe instruction-following instability, with 40.6\% of tasks failing due to invalid output generation. Overall, the results show that while enhanced reasoning capabilities significantly reduce syntax and execution errors, the demands of long-horizon, constraint-heavy planning remain a persistent challenge even for state-of-the-art systems.

\subsection{Leveling + distractor noise mechanics}
Table \ref{tab:level_noise} presents the results for level 10 difficulty tasks (level 9, plus leveling mechanics and distractor noise items), evaluated on top-performing models. Grok-4 demonstrates a significant lead over all other models, with only Gemini 2.5 Pro and GPT-5 managing to solve a subset of the hardest tasks. Notably, Grok-4's performance remains consistently high across all difficulty levels, showing drop in success rate with addition of leveling and score with addition of noise items. The performance of GPT-5 decreases with the addition of leveling mechanics but remains unaffected by the inclusion of additional noise items. Notably, GPT-5 and Grok-4 are the only two models that substantially increase their reasoning length as task difficulty rises. The results of Grok-4 also show that the tasks are solvable within the 20-35k output tokens. The details of Grok-4’s architecture and training remain undisclosed, though its impressive performance may be attributed to the reported large-scale reinforcement learning applied during post-training.

\subsection{Pass@k metric}
While recent findings suggest that reinforcement learning with verifiable rewards (RLVR) may not consistently improve over the base model's pass@k performance when k is sufficiently large \cite{yue2025doesreinforcementlearningreally}, our results indicate that this conclusion may be task-dependent. In the context of our planning benchmark, particularly at difficulty levels 1 and 2, we observed that, even after 200 attempts, the base models Qwen3-8B and Qwen3-32B were unable to match the performance of their reasoning-enabled counterparts, which achieved higher pass rates with just 10 attempts Table~\ref{tab:qwen-passatk}. Since these lower-difficulty tasks do not demand long reasoning chains, as evidenced by the success of other non-reasoning models on them,this gap suggests that RLVR can provide tangible benefits in planning scenarios where structured reasoning is essential.

\begin{table}[t]
\centering
\begin{tabular}{|l|c|c|c|c|c|}
\hline
\textbf{Model}    & \textbf{Diff} & \textbf{k} & \textbf{pass@k} & \textbf{Mean Win} \\
\hline
Qwen3-8B   & 1 & 200 & 45.0\% & 11.8\% \\
Qwen3-8B (t)   & 1 & 10  & 65.0\% & 30.5\% \\
Qwen3-32B  & 2 & 200 & 30.0\% & 0.6\%  \\
Qwen3-32B (t)  & 2 & 10 & 75.0\% & 20.0\% \\
\hline
\end{tabular}
\caption{Performance of thinking (t) and non-thinking Qwen3 Models using pass@k metric. Results suggest that the RLVR approach noticeably improves the results and may be task-dependent.}
\label{tab:qwen-passatk}
\end{table}

\section{Related work}
Planning is a fundamental capability for LLMs and LLM-based agents, and its reliable evaluation requires benchmarks that capture multi-step reasoning, state transitions, constraints, and long-horizon decision making \cite{yehudai2025surveyevaluationllmbasedagents,wei2025plangenllmsmodernsurveyllm,li2025planetcollectionbenchmarksevaluating}
A variety of benchmarks have been proposed to evaluate the planning capabilities of LLM-based agents across symbolic, embodied, and real-world domains. PlanBench~\cite{NEURIPS2023_7a92bcde} evaluates classical planning by testing reasoning about actions and state transitions beyond commonsense recall. ALFWorld~\cite{shridhar2021alfworldaligningtextembodied} studies multi-step household planning by aligning abstract language plans with embodied execution, while ScienceWorld~\cite{wang2022scienceworldagentsmarter5th} focuses on procedural scientific planning through interactive experiments rather than question answering. Natural Plan~\cite{zheng_natural_2024} evaluates realistic planning in natural language across trip planning, meeting scheduling, and calendar coordination tasks, and TravelPlanner~\cite{xie_travelplanner_2024} extends this setting with large-scale data, tool use, and complex real-world constraints. LogiPlan~\cite{cai2025logiplanstructuredbenchmarklogical} targets logical and relational planning by requiring models to generate and verify structured dependency graphs. ROBOTOUILLE~\cite{gonzalezpumariega2025robotouilleasynchronousplanningbenchmark} introduces asynchronous planning with time delays and long-horizon dependencies in a cooking domain. Plancraft~\cite{dagan_plancraft_2025} presents a multimodal Minecraft-based benchmark for evaluating hierarchical planning, resource management, and feasibility recognition, including unsolvable tasks.




\section{Conclusions}
This work introduced HeroBench, a benchmark designed to stress-test long-horizon planning and structured reasoning in LLMs and LRMs under tightly constrained conditions. Unlike prior planning benchmarks that emphasize abstract symbolic domains, lack difficulty and depth or rely on continuous interaction and feedback, HeroBench evaluates one-shot, end-to-end planning in a rich and complex environment with many interdependent rules, constraints, and dependencies, making it well suited to challenging the latest generation of models.

Our extensive evaluation of LLMs across a range of sizes and model families shows that RL-based post-training enables substantial gains over base models, yet performance still degrades as task horizon, dependency depth, and numeric feasibility constraints increase. Even the strongest reasoning models fail to reliably solve the hardest tasks, despite producing long and detailed reasoning traces, while smaller and open-source models exhibit very low reliability even on comparatively simple tasks.

Overall, HeroBench provides a challenging and transparent testbed that exposes limitations of current LLMs in long-horizon, constraint-heavy planning, offering a clear target for future advances in autonomous reasoning systems.
\section{Future work}

 Although the current set of tasks is designed to evaluate existing systems without overwhelming them with excessive complexity, the environment natively supports multi-agent dynamics, seamless transition to open-ended play, and integration of visual modalities. Future extensions may include multi-agent manipulation, collaboration and competition dynamics, additional in-game mechanics, stochasticity in the tasks and a natural extension of HeroBench into a RL environment.

\bibliographystyle{named}
\bibliography{ijcai26}

\clearpage
\appendix

\onecolumn

\section{Model Inference Configuration}
All experiments using Openrouter API were conducted using default hyperparameters:  
\texttt{temperature = 1.0}, \texttt{top\_p = 1.0}, \texttt{top\_k = 0}, \texttt{frequency\_penalty = 0.0}, \texttt{presence\_penalty = 0.0}, \texttt{repetition\_penalty = 1.0}, \texttt{min\_p = 0.0}, \texttt{top\_a = 0.0}.

The Qwen 3 models were run with the same hyperparameters on a system equipped with 2× A100 GPUs (80GB VRAM each).

\section{Multi-Agent systems architecture}
\label{app:multi_agent_descr}
\begin{figure}[ht]
  \centering{
  \includegraphics[width=0.8\linewidth]{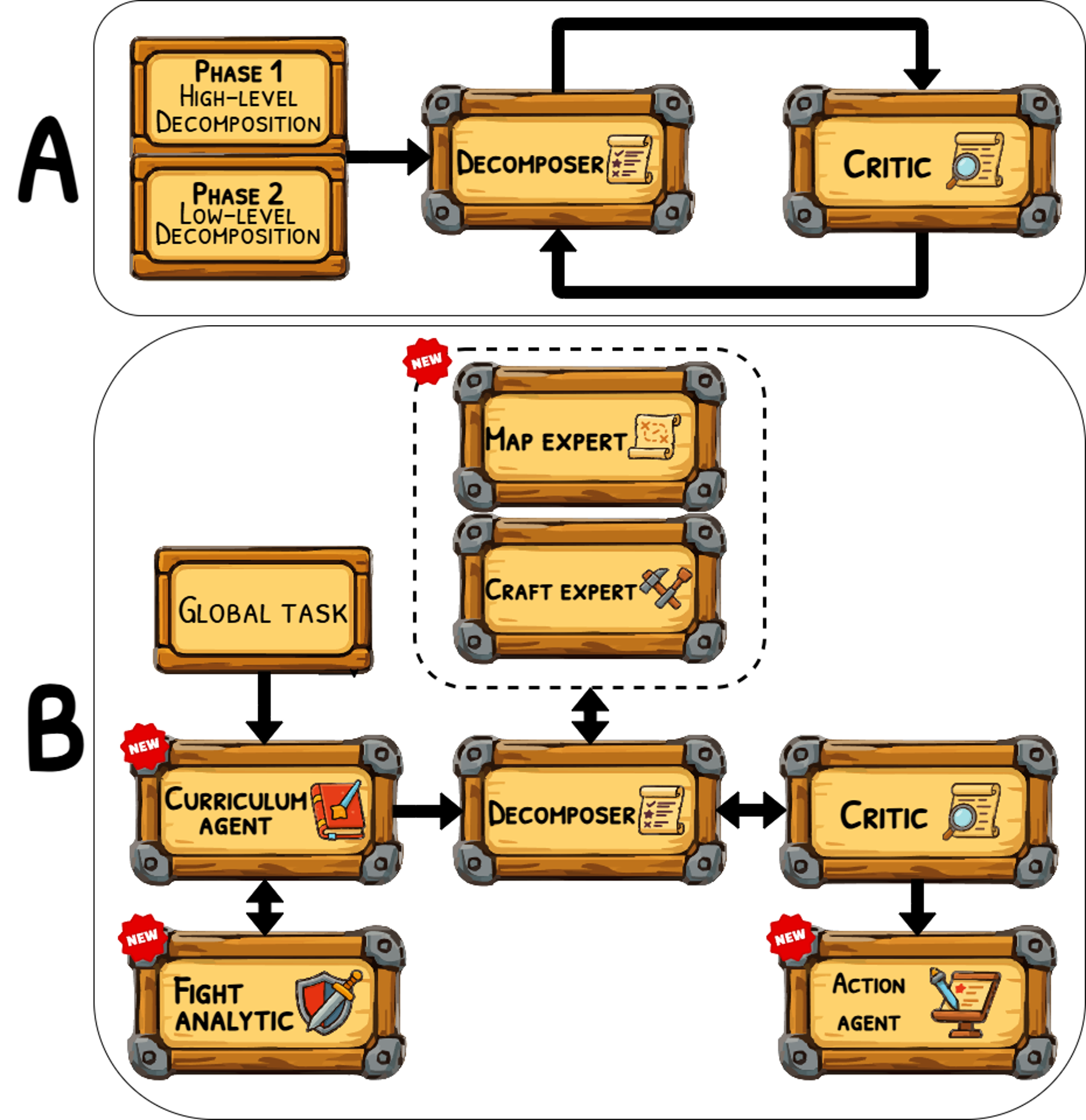}
  }
  \caption{Two agent architectures were evaluated in the benchmark. A: A-1, operates in two phases: in the first phase, it generates a high-level plan; in the second, it decomposes this plan into executable actions. B: Agentic system A-2, is a modification of A-1 based on the idea of assigning each agent a single, one-bite task.}
  \label{fig:agents_arch}
\end{figure}

A-1 represents the first version of the multi-agent system developed for HeroBench. The A-1 architecture employs a pair of agents: a decomposer/action agent and a critic agent. This agent pair is used to perform two-level task decomposition. The architecture of A-1 is shown in Fig.~\ref{fig:agents_arch}A.

At the first level, the task is divided into high-level subtasks, which are used to form a high-level plan. Once the plan is verified by the critic agent, it is passed to the second stage of decomposition. In this stage, each item in the high-level plan is further broken down into basic subtasks, resulting in low-level steps (actions). The collection of all actions constitutes the final executable plan, which is carried out in the environment through calls to the corresponding environment API functions. To prevent infinite loops, the system imposes limits on the number of allowed decomposition attempts. For decomposition, the agent relies on the current environment state and automatically generated descriptions of the tasks or subtasks target objects.

A-2 is a more complex multi-agent system designed to evaluate a deeper, linear hierarchy of cooperating agents. The main idea is to allow agents to solve simple, one-bite tasks individually while assisting one another. This system builds upon the TaskGen agentic framework~\cite{tan_taskgen_2024}, extending the initial A-1 architecture (Fig.~\ref{fig:agents_arch}B).

Several new agents were introduced in A-2. First, a curriculum agent formulates high-level plans based on the global task and the current state of the characters. An optional fight analytic agent estimates the outcomes of combat encounters, taking into account character statuses, equipment, and potential crafted items. Additional subagents were also integrated into the decomposer agent: a map expert and a craft expert, both capable of acquiring world knowledge and answering the decomposer's queries. Finally, an action agent was added to generate the executable Python code to complete the task.

Both A-1 and A-2 are also capable of solving open-ended tasks by incorporating feedback from the environment.

\subsection{Multi-Agent systems performance}
To evaluate the multi-agent systems, tasks with difficulty levels 2 and 3 were selected. GPT-4.1-mini was chosen as the baseline model due to its accessibility, high speed, and reasonably strong performance. The experimental results (Table~\ref{tab:agents_difficulty}) show that the A-1 multi-agent system achieved a higher success rate than the baseline model. However, the performance of A-2 was lower than that of the baseline, presumably due to its more complex architecture and prompt overengineering. Smaller models were unable to effectively process the context provided by the subagents and exhibited hallucinations during plan and subtask generation. These findings indirectly suggest that while multi-agent systems can better maintain problem-solving capabilities at higher task complexities, they require very careful design - particularly with regard to task complexity and prompt size.

\begin{table}[ht]
\centering
\begin{tabular}{|l|c|c|}
\hline
\textbf{Agent/Model} & \textbf{Difficulty 2} & \textbf{Difficulty 3} \\
\hline
A-1          & 65\% & 60\% \\
\hline
A-2          & 35\% & 10\% \\
\hline
GPT-4.1-mini & 45\% & 15\% \\
\hline
\end{tabular}
\caption{A-1, A-2 and GPT-4.1-mini success rate comparison. The results indicate that the simple decomposer-critic loop for the small models outperformed a modified decomposition architecture with one-bite tasks.}
\label{tab:agents_difficulty}
\end{table}

\section{Prompt Example}
\label{app:prompt}
This section contains example prompt from HeroBench dataset.

\lstdefinestyle{promptstyle}{
  basicstyle=\ttfamily\small,
  breaklines=true,
  breakatwhitespace=false,
  columns=fullflexible,
  keepspaces=true,
  showstringspaces=false,
  upquote=true
}

\newtcolorbox{promptbox}{
  colback=gray!7!white,
  colframe=black!60,
  boxrule=0.7pt,
  sharp corners,
  enhanced,
  breakable,
  title=Prompt Example,
}

\begin{promptbox}
\begin{lstlisting}[style=promptstyle]
I will provide you with the information about game environment and a task you need to accomplish.

Rules for items:
Weapons have attack stats that will provide the damage type and damage value, for example {'name':
'attack_air', 'value': 8} means air damage type and 8 damage per turn. Armor and jewelry can have hp
stat, that will add hp number to the players hp {'name': 'hp', 'value': 20}, damage amplifications
stat in % {'name': 'dmg_fire', 'value': 3} and resistance stat in % {'name': 'res_air', 'value': 3}.
The damage amplification will increase the attack of your current weapon by the stated value if you
have matching attack type weapon.

Rules for Fighting Monsters:
To fight a monster you need to be on the same location as monster and perform fight action.
Combat follows a turn-based system where you attack first, followed by the monster. There are
different damage types: fire, earth, water, and air. If a character or monster has resistance
matching the attacker's damage type, the damage is either reduced or amplified based on the
resistance percentage.
The calculations of damage follow these formulas:
character_damage = base_attack + dmg_boost - res_reduction
monster_damage = base_attack - res_reduction
res_reduction = base_attack * (resist_percentage / 100)
dmg_boost = base_attack * (dmg_percentage / 100)

The damage is calculated from every non 0 attack type the monster or character has. After each turn,
the health points of the character or monster decrease by the calculated damage. The fight is
automatic and continues until the health points of one participant drop to 0.
The items acquired from monsters drop after you defeat them in a fight. Assume that drop rates from
monsters are 100%.
Your health and monsters health is restored to full after each fight.

Rules for crafting:
Each gathering action provides you one resource.
To craft an item you should be on the appropriate location like crafting station for the item you
want to craft. Your crafting level should not be less than required.

You can perform the following actions:

move(character_name: str, x: int, y: int)
fight(character_name: str)
equip(character_name: str, slot: str, item_name: str, quantity: int = 1)
unequip(character_name: str, slot: str, quantity: int = 1)
Item slot. Allowed values: 'weapon', 'shield', 'helmet', 'body_armor', 'leg_armor', 'boots',
         'ring1', 'ring2', 'amulet', 'artifact1', 'artifact2', 'artifact3', 'consumable1' or
         'consumable2'.

gather(character_name: str)
craft(character_name: str, item_name: str, quantity: int)
Use codes of items in functions like 'copper_dagger'. You can only use these functions and for loops
in your answer, nothing else. Do not include parameter names when calling functions.
If there is an item in the slot, you should unequip it before equipping a different item.
{'Monsters': [{'name': 'Lich', 'code': 'lich', 'level': 30, 'hp': 1500, 'attack_fire': 60,
'attack_earth': 60, 'attack_water': 0, 'attack_air': 0, 'res_fire': 24, 'res_earth': 24,
'res_water': 18, 'res_air': 18, 'min_gold': 0, 'max_gold': 15, 'drops': [{'code': 'life_crystal',
'rate': 3000, 'min_quantity': 1, 'max_quantity': 1}, {'code': 'lich_crown', 'rate': 800,
'min_quantity': 1, 'max_quantity': 1}]}, {'name': 'Ogre', 'code': 'ogre', 'level': 20, 'hp': 680,
'attack_fire': 0, 'attack_earth': 80, 'attack_water': 0, 'attack_air': 0, 'res_fire': -20,
'res_earth': 30, 'res_water': 0, 'res_air': 0, 'min_gold': 0, 'max_gold': 3, 'drops': [{'code':
'ogre_eye', 'rate': 20, 'min_quantity': 1, 'max_quantity': 1}, {'code': 'ogre_skin', 'rate': 20,
'min_quantity': 1, 'max_quantity': 1}]}], 'Craftable items': [{'name': 'Dreadful Ring', 'code':
'dreadful_ring', 'level': 20, 'type': 'ring', 'subtype': '', 'description': '', 'effects': [{'name':
'hp', 'value': 20}, {'name': 'dmg_earth', 'value': 11}, {'name': 'dmg_water', 'value': 11}],
'craft': {'skill': 'jewelrycrafting', 'level': 20, 'items': [{'code': 'steel', 'quantity': 7},
{'code': 'ogre_eye', 'quantity': 4}, {'code': 'ogre_skin', 'quantity': 3}, {'code':
'jasper_crystal', 'quantity': 1}], 'quantity': 1}}, {'name': 'Steel', 'code': 'steel', 'level': 20,
'type': 'resource', 'subtype': 'alloy', 'description': '', 'effects': [], 'craft': {'skill':
'mining', 'level': 20, 'items': [{'code': 'iron_ore', 'quantity': 3}, {'code': 'coal', 'quantity':
5}], 'quantity': 1}}], 'Resources': [{'name': 'Iron Ore', 'code': 'iron_ore', 'level': 10, 'type':
'resource', 'subtype': 'mining', 'description': '', 'effects': [], 'craft': None, 'sources':
['iron_rocks']}, {'name': 'Coal', 'code': 'coal', 'level': 20, 'type': 'resource', 'subtype':
'mining', 'description': '', 'effects': [], 'craft': None, 'sources': ['coal_rocks']}, {'name':
'Ogre Eye', 'code': 'ogre_eye', 'level': 20, 'type': 'resource', 'subtype': 'mob', 'description':
'', 'effects': [], 'craft': None}, {'name': 'Ogre Skin', 'code': 'ogre_skin', 'level': 20, 'type':
'resource', 'subtype': 'mob', 'description': '', 'effects': [], 'craft': None}, {'name': 'Jasper
Crystal', 'code': 'jasper_crystal', 'level': 15, 'type': 'resource', 'subtype': 'mining',
'description': '', 'effects': [], 'craft': None, 'sources': ['jasper_rocks']}], 'Locations':
[{'name': 'Forest', 'skin': 'forest_ironore2', 'x': 1, 'y': 7, 'content': {'type': 'resource',
'code': 'iron_rocks'}}, {'name': 'Forest', 'skin': 'forest_coal1', 'x': 1, 'y': 6, 'content':
{'type': 'resource', 'code': 'coal_rocks'}}, {'name': 'Forest', 'skin': 'forest_ogre1', 'x': -5,
'y': -5, 'content': {'type': 'monster', 'code': 'ogre'}}, {'name': 'Forest', 'skin': 'forest_ogre2',
'x': -5, 'y': -4, 'content': {'type': 'monster', 'code': 'ogre'}}, {'name': 'Forest', 'skin':
'forest_2', 'x': 0, 'y': 6, 'content': {'type': 'resource', 'code': 'jasper_rocks'}}, {'name':
'Graveyard', 'skin': 'forest_skeleton5', 'x': 9, 'y': 7, 'content': {'type': 'monster', 'code':
'lich'}}, {'name': 'Forest (Forge)', 'skin': 'forest_miningstation1', 'x': 1, 'y': 5, 'content':
{'type': 'workshop', 'code': 'mining'}}, {'name': 'City', 'skin': 'forest_jewelrycrafting1', 'x': 1,
'y': 3, 'content': {'type': 'workshop', 'code': 'jewelrycrafting'}}], 'Items stats': []}
Character Stats:
{'name': 'Hero', 'skin': 'men1', 'level': 30, 'xp': 0, 'max_xp': 150, 'mining_level': 20,
'mining_xp': 0, 'mining_max_xp': 150, 'woodcutting_level': 1, 'woodcutting_xp': 0,
'woodcutting_max_xp': 150, 'fishing_level': 1, 'fishing_xp': 0, 'fishing_max_xp': 150,
'weaponcrafting_level': 1, 'weaponcrafting_xp': 0, 'weaponcrafting_max_xp': 150,
'gearcrafting_level': 1, 'gearcrafting_xp': 0, 'gearcrafting_max_xp': 150, 'jewelrycrafting_level':
20, 'jewelrycrafting_xp': 0, 'jewelrycrafting_max_xp': 150, 'cooking_level': 1, 'cooking_xp': 0,
'cooking_max_xp': 150, 'hp': 965, 'attack_fire': 0, 'attack_earth': 20, 'attack_water': 60,
'attack_air': 0, 'dmg_fire': 25, 'dmg_earth': 11, 'dmg_water': 103, 'dmg_air': 5, 'res_fire': 19,
'res_earth': 17, 'res_water': 24, 'res_air': 17, 'x': 0, 'y': 0, 'weapon_slot':
'greater_dreadful_staff', 'shield_slot': 'steel_shield', 'helmet_slot': 'gold_helm',
'body_armor_slot': 'lizard_skin_armor', 'leg_armor_slot': 'obsidian_legs_armor', 'boots_slot':
'gold_boots', 'ring1_slot': 'dreadful_ring', 'ring2_slot': '', 'amulet_slot': 'sapphire_amulet',
'artifact1_slot': '', 'artifact2_slot': '', 'artifact3_slot': '', 'consumable1_slot': '',
'consumable1_slot_quantity': 0, 'consumable2_slot': '', 'consumable2_slot_quantity': 0, 'inventory':
[]}
Character stats include bonuses from equipped gear.

Your task is to kill 1 Lich. Make sure you can defeat it.
You may need to craft and equip several items to do it. Make sure you have all resources for
crafting. Do not make assumptions, calculate everything.
Think and write a sequence of actions to do it.
End your answer with the following: 'Final answer: <python code to solve the task>'
\end{lstlisting}
\end{promptbox}

\section{Additional Charts}
\label{app:additional_results}
\begin{figure*}[!h]
  \centering
  \includegraphics[width=0.9\textwidth]{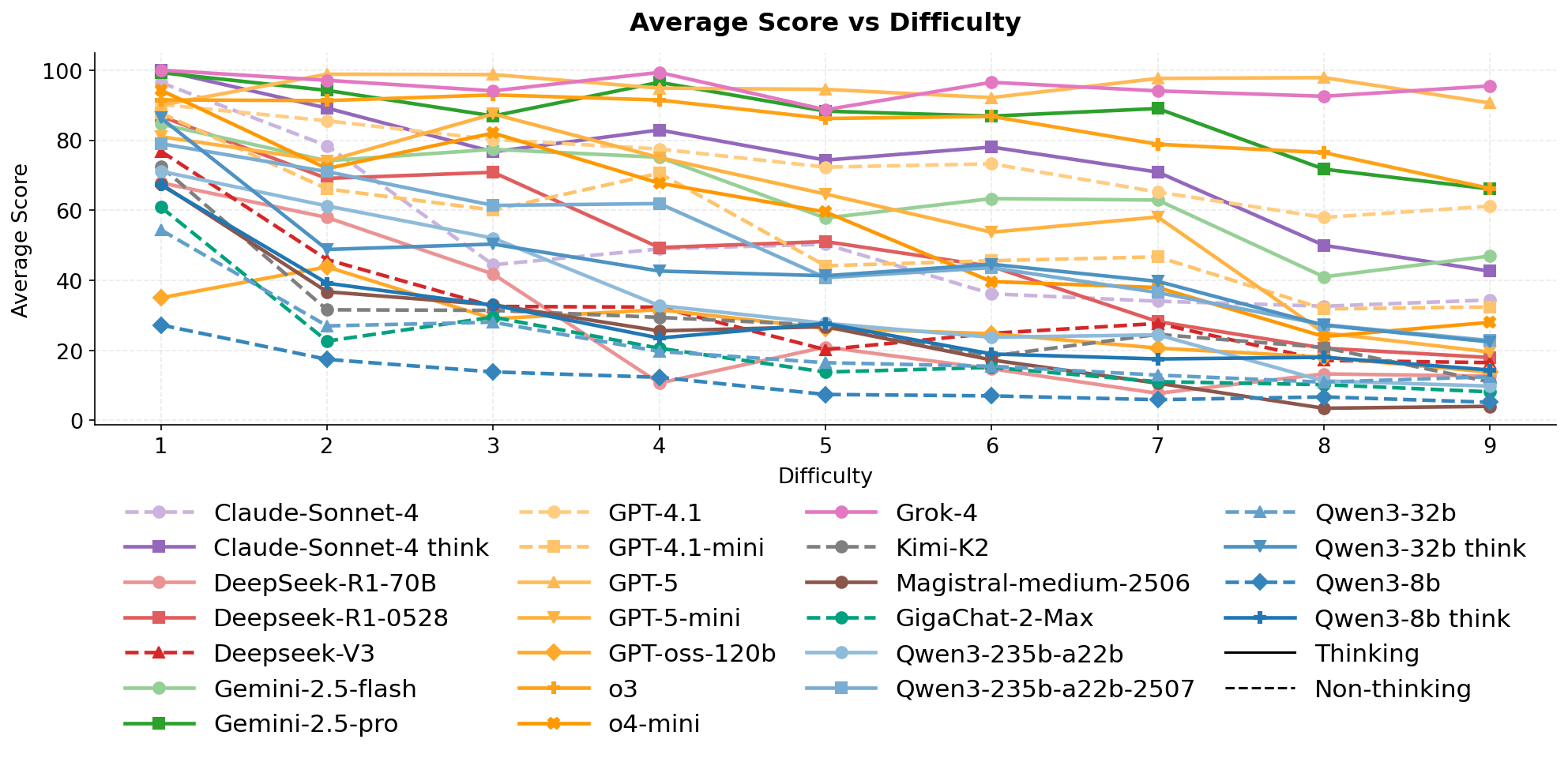}
  \caption{Score of LLMs across nine base-task difficulty levels. Solid lines correspond to reasoning-enabled (thinking)
models, while dashed lines represent standard (non-thinking) variants.}
  \label{fig:results5}
\end{figure*}

\begin{figure*}[!h]
  \centering
  \includegraphics[width=0.9\textwidth]{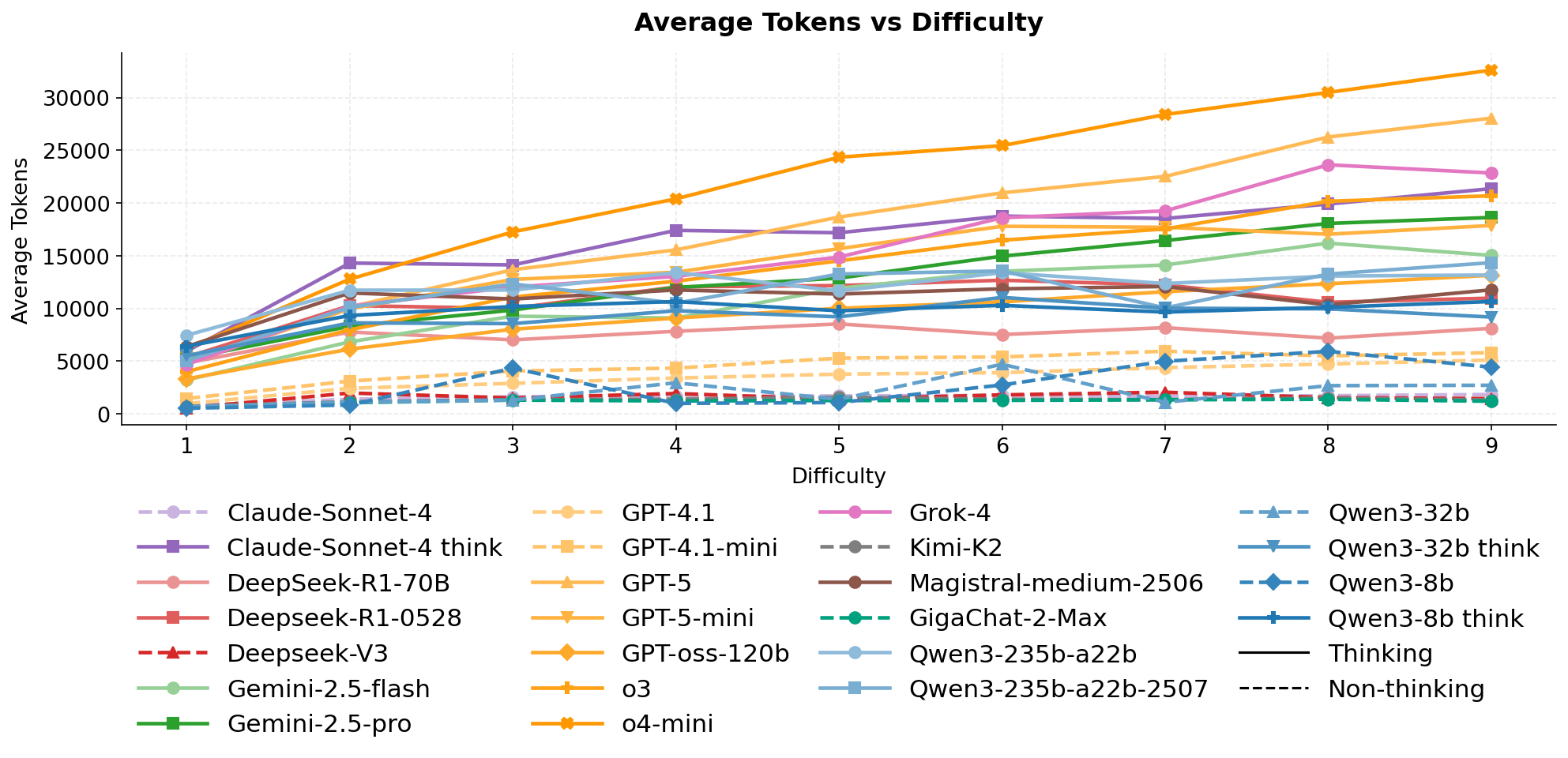}
  \caption{Token use of LLMs across nine base-task difficulty levels. Solid lines correspond to reasoning-enabled (thinking)
models, while dashed lines represent standard (non-thinking) variants.}
  \label{fig:results6}
\end{figure*}

\begin{figure*}[!h]
  \centering
  \includegraphics[width=0.9\textwidth]{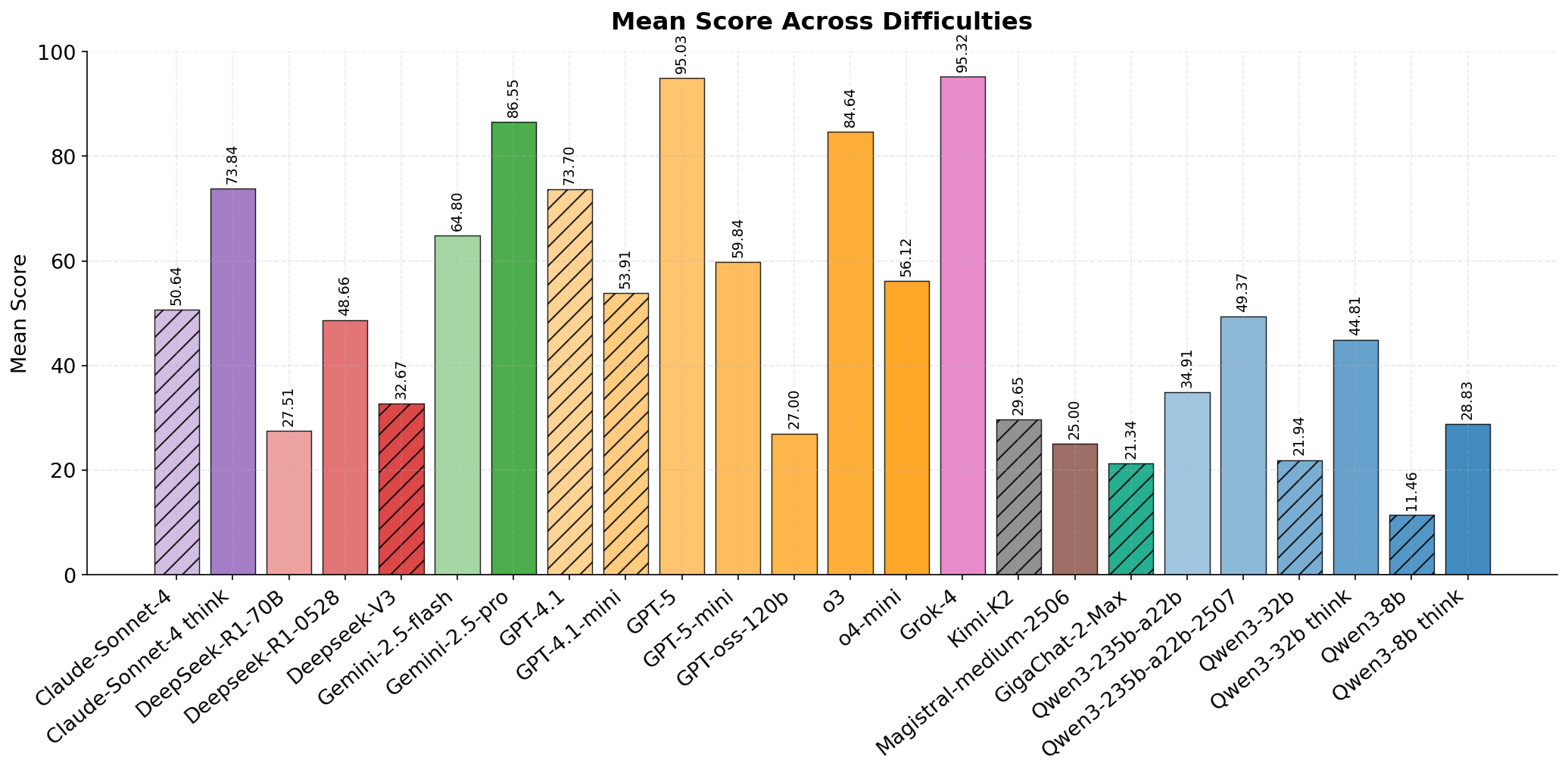}
  \caption{Mean score of LLMs over nine base-task difficulty levels. Solid bars correspond to reasoning-enabled (thinking)
models, while hatched bars represent standard (non-thinking) variants.}
  \label{fig:results7}
\end{figure*}

\begin{figure*}[!h]
  \centering
  \includegraphics[width=0.9\textwidth]{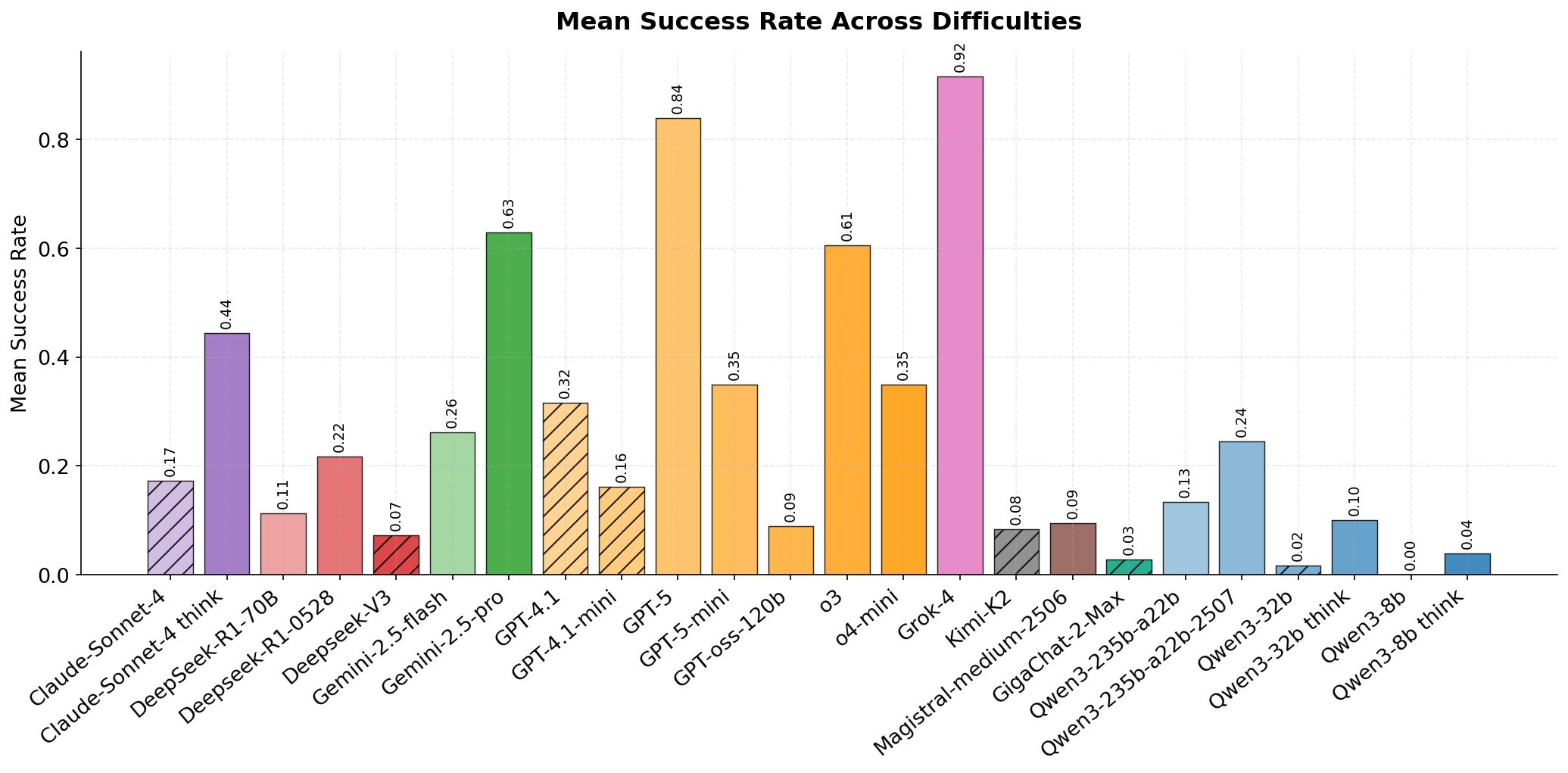}
  \caption{Mean success rate of LLMs over nine base-task difficulty levels. Solid bars correspond to reasoning-enabled (thinking)
models, while hatched bars represent standard (non-thinking) variants.}
  \label{fig:results8}
\end{figure*}

\begin{figure*}[!h]
  \centering
  \includegraphics[width=0.9\textwidth]{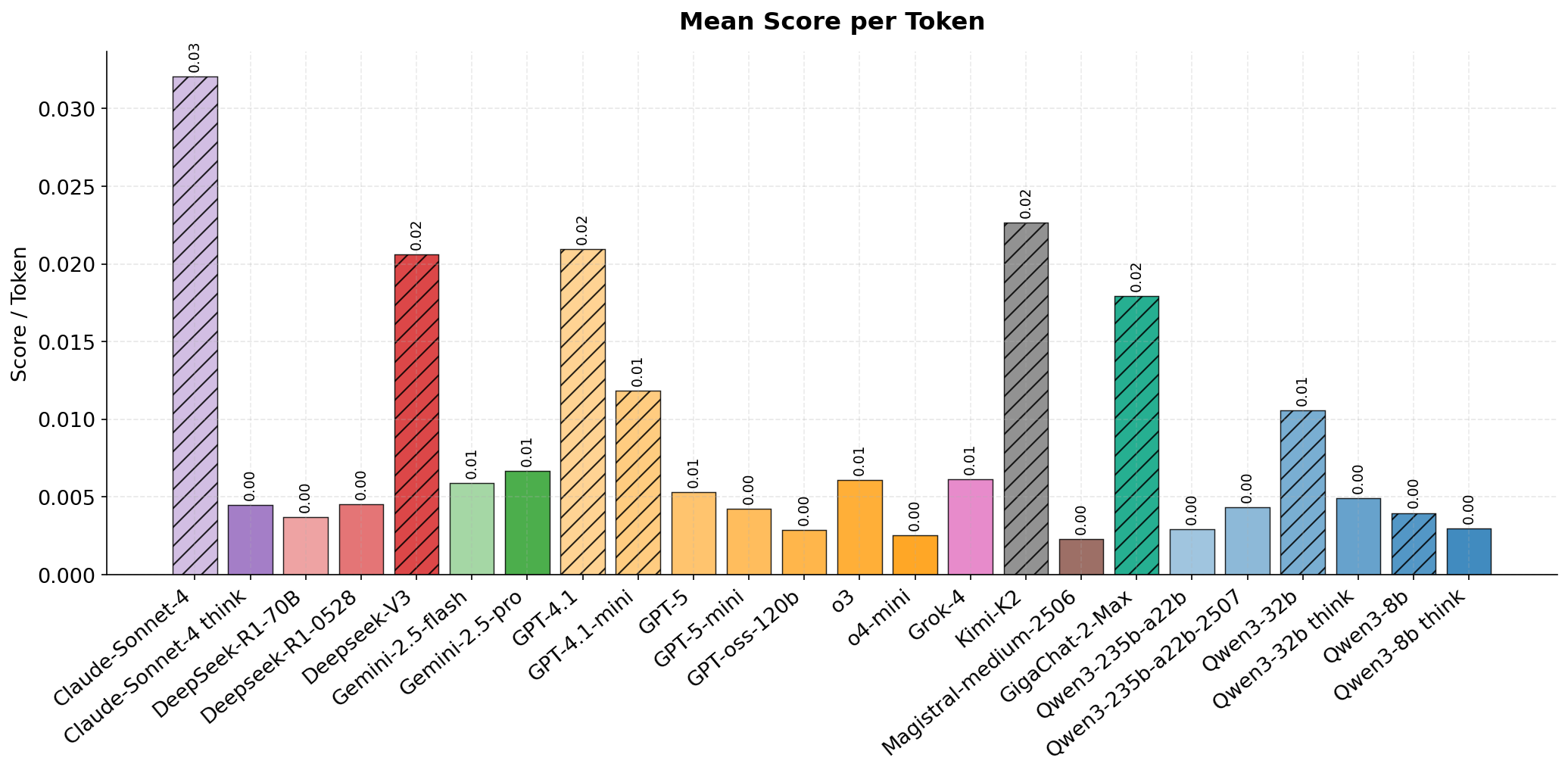}
  \caption{Mean score per token of LLMs over nine base-task difficulty levels. Solid bars correspond to reasoning-enabled (thinking) models, while hatched bars represent standard (non-thinking) variants.}
  \label{fig:results9}
\end{figure*}

\begin{figure*}[!h]
  \centering
  \includegraphics[width=0.9\textwidth]{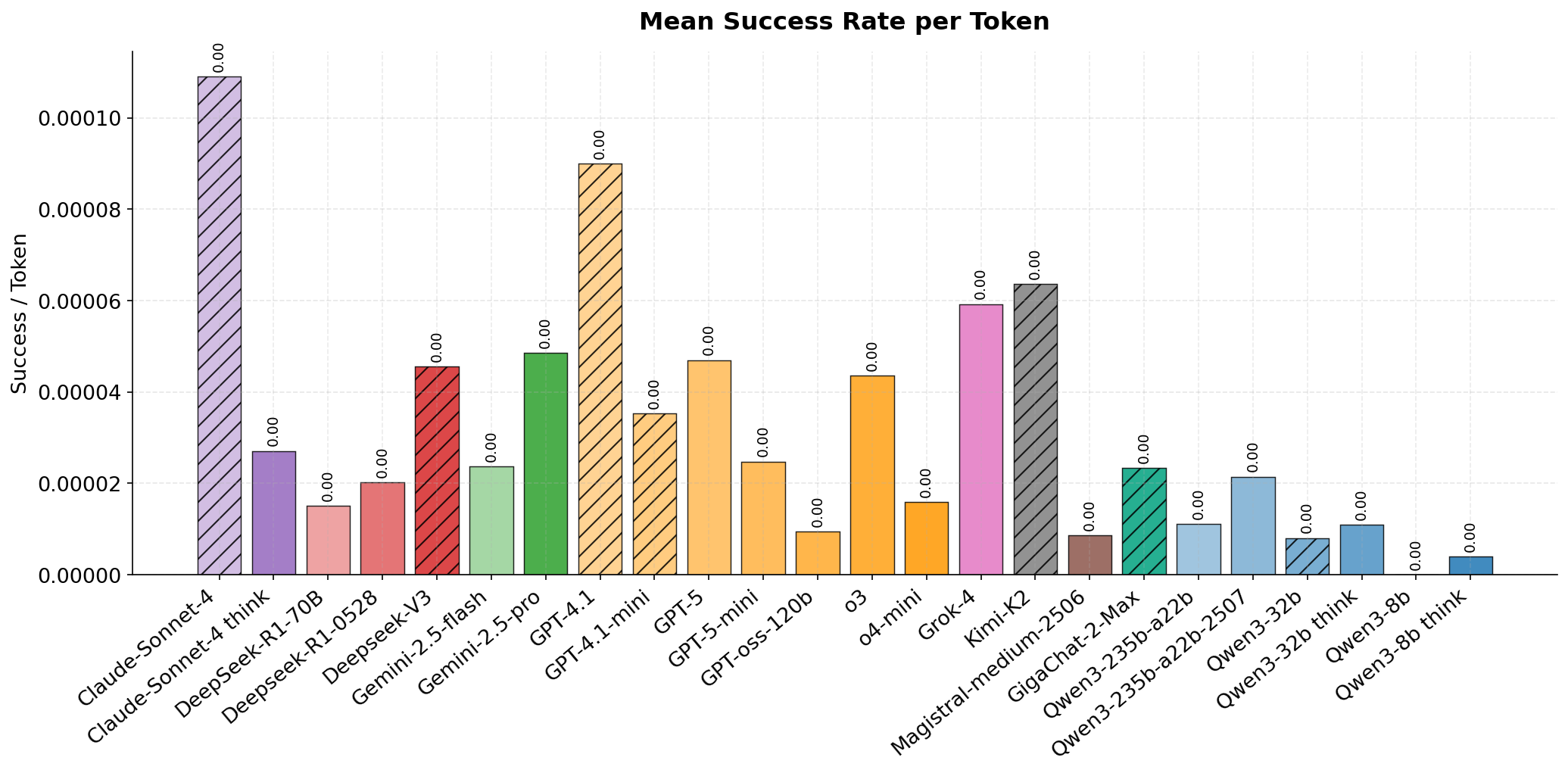}
  \caption{Mean success rate per token of LLMs over nine base-task difficulty levels. Solid bars correspond to reasoning-enabled (thinking) models, while hatched bars represent standard (non-thinking) variants.}
  \label{fig:results10}
\end{figure*}

\bigskip

\end{document}